\documentclass[letterpaper, 10 pt, conference]{ieeeconf}  

\IEEEoverridecommandlockouts                              

\usepackage{amsmath,amsfonts}
\usepackage{algorithmic}
\usepackage{array}
\usepackage{kotex}
\usepackage{xcolor}
\usepackage{ulem}
\usepackage{textcomp}
\usepackage{stfloats}
\usepackage{url}
\usepackage{verbatim}
\usepackage{graphicx}
\usepackage{multirow}
\usepackage{hhline}
\usepackage{adjustbox}
\usepackage{booktabs}
\usepackage{makecell}
\usepackage{bbm}
\usepackage{xcolor,colortbl}
\usepackage{varwidth}
\usepackage{threeparttable} 
\usepackage{dsfont}
\usepackage{soul} 

\newcommand{\Idf}{\mathds{1}} 
\newcommand{\ccgray}{\cellcolor{gray!15}}

\title{\LARGE \bf
Boosting Cross-spectral Unsupervised Domain Adaptation for Thermal Semantic Segmentation
}

\author{Seokjun Kwon$^{1*}$, Jeongmin Shin$^{1*}$, Namil Kim$^{2}$, Soonmin Hwang$^{3}$, Yukyung Choi$^{1}$\textsuperscript{\textdagger}
    \thanks{This work was partly supported by the IITP(Institute of Information \& Coummunications Technology Planning \& Evaluation)-ICAN(ICT Challenge and Advanced Network of HRD)(IITP-2025-RS-2022-00156345, 50\%) and ITRC(Information Technology Research Center)(IITP-2025-RS-2024-00437494, 25\%) grant funded by the Korea government(MSIT). This research was partly supported by Unmanned Vehicles Core Technology Research and Development Program through the National Research Foundation of Korea (NRF) and Unmanned Vehicle Advanced Research Center (UVARC) funded by the Ministry of Science and ICT, the Republic of Korea (NRF-2023M3C1C1A01098408, 25\%).}%
    \thanks{$^{1}$Seokjun Kwon, Jeongmin Shin, and Yukyung Choi are with the Sejong University, South Korea {\tt\small \{sjkwon, jmshin, ykchoi\}@rcv.sejong.ac.kr}}%
    \thanks{$^{2}$ Namil Kim is with NAVER LABS, South Korea {\tt\small namil.kim@naverlabs.com}}%
    \thanks{$^{3}$ Soonmin Hwang is with the Department of Automotive Engineering, Hanyang University, South Korea {\tt\small soonminh@hanyang.ac.kr}}%
    \thanks{$^{*}$: Equal Contribution, \textsuperscript{\textdagger}: Corresponding Author}
}

\begin{document}

\maketitle
\thispagestyle{empty}
\pagestyle{empty}

\begin{abstract}

In autonomous driving, thermal image semantic segmentation has emerged as a critical research area, owing to its ability to provide robust scene understanding under adverse visual conditions.
In particular, unsupervised domain adaptation (UDA) for thermal image segmentation can be an efficient solution to address the lack of labeled thermal datasets.
Nevertheless, since these methods do not effectively utilize the complementary information between RGB and thermal images, they significantly decrease performance during domain adaptation.
In this paper, we present a comprehensive study on cross-spectral UDA for thermal image semantic segmentation. We first propose a novel masked mutual learning strategy that promotes complementary information exchange by selectively transferring results between each spectral model while masking out uncertain regions.
Additionally, we introduce a novel prototypical self-supervised loss designed to enhance the performance of the thermal segmentation model in nighttime scenarios. 
This approach addresses the limitations of RGB pre-trained networks, which cannot effectively transfer knowledge under low illumination due to the inherent constraints of RGB sensors.
In experiments, our method achieves higher performance over previous UDA methods and comparable performance to state-of-the-art supervised methods.

\end{abstract}

\section{INTRODUCTION}

\begin{figure}[!t]
    \centering
    \includegraphics[width=0.74\linewidth]{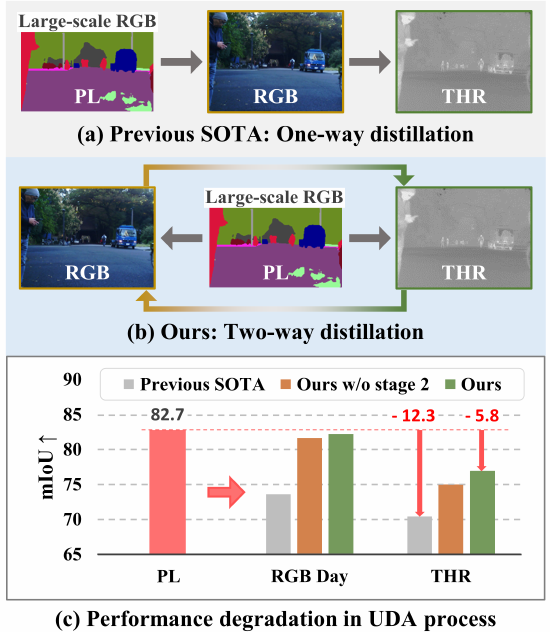}
    \caption{\label{intro} To handle the labeled data scarcity problem in the thermal domain, we employ a pre-trained model on a large-scale RGB dataset~\cite{CityScapes} to train student networks. (a) Previous SOTA method~\cite{MS-UDA} performs one-way distillation from PL to THR, disregarding the characteristics of each spectral domain. (b) Our approach adopts two-way distillation, which appropriately transfers the complementary knowledge of each spectral domain.
    (c) To validate the potential of the distillation processes, we evaluate the performance of the teacher (i.e., PL) and each spectral student network. Our final method outperforms the previous SOTA method in both RGB and thermal domains despite leveraging the same pseudo-labels at the training phase. Due to the limitations of the RGB sensor at nighttime, previous SOTA and our method leverage only daytime RGB images for the distillation process (i.e., RGB Day).
    (PL: pseudo-labels, THR: thermal)}
    \label{fig:intro}
    \vspace*{-\baselineskip}
\end{figure}

In recent years, there has been a significant increase in research on robust semantic segmentation techniques for challenging environments. This surge is driven by the critical need for reliable performance in autonomous driving, as it directly impacts human safety. Nevertheless, previous approaches~\cite{DeepV3, HRNet, robot2} relying solely on visual cues from RGB sensors often struggle in adverse scenarios such as low-light conditions, dense fog, and heavy rain. To tackle this issue, thermal sensors that capture the heat signatures of objects have been extensively utilized for achieving reliable semantic segmentation in challenging conditions~\cite{EC-CNN, MCNet}.

However, several challenges must be overcome in thermal image semantic segmentation to improve performance gain. Specifically, in these methods training the model is impeded by the lack of large-scale datasets, where pixel-level annotations are labor-intensive and costly to obtain. Moreover, thermal images often suffer from low quality such as textureless and low resolution, which could harm model training and performance gain despite the advantages of thermal cameras in low-illumination conditions.
To tackle these issues, a few unsupervised domain adaptation (UDA) methods~\cite{MS-UDA} have emerged in cross-spectral domains. The main idea of MS-UDA{~\cite{MS-UDA}} is to leverage the knowledge learned from a large-scale RGB dataset with segmentation labels, which serves as the source domain, and transfer this knowledge to the thermal image domain where manual labels are not available, as illustrated in Fig.{~\ref{fig:intro}}-(a).

However, one limitation of MS-UDA lies in its reliance on a one-way knowledge distillation process that involves training an RGB student network using a pre-trained teacher model on the large-scale RGB dataset. The outputs of the RGB network are then used as training labels for the thermal network. Since the student network often inherits equal or lower performance relative to the teacher, this one-way distillation strategy restricts the model's potential for achieving higher performance, as depicted in Fig.{~\ref{fig:intro}}-(c) In addition, the knowledge distillation process of MS-UDA relies solely on distilling the final predicted values across spectral domains. This disregard for domain-specific features inherent in RGB and thermal data impedes successful cross-spectral adaptation.
Moreover, MS-UDA generates day-to-night synthetic thermal images for domain generalization in thermal images due to the inaccuracy of pseudo-labels from the RGB network in low-light conditions. These fake night images, however, might not fully reflect the whole distribution of real nighttime scenes, leading to limited performance improvements.

In this paper, we present novel learning and loss strategies that significantly improve the performance of cross-spectral unsupervised domain adaptation framework for thermal image semantic segmentation. We first propose a novel masked mutual learning that distills the knowledge learned from a large-scale RGB dataset in a two-way, as depicted in Fig.~\ref{fig:intro}-(b). Our proposed strategy encourages the RGB and thermal student network to learn complementary information from each spectral knowledge by filtering the uncertain pixels.
Furthermore, our novel prototypical self-supervised loss enables training of the model when the pseudo-labels provided by the teacher network are unreliable due to challenging conditions such as poor illumination at night. 

\section{RELATED WORK}

\subsection{Unsupervised Domain Adaptation}
Unsupervised Domain Adaptation (UDA) transfers knowledge acquired from a large-scale source domain to a target domain where labeled data are scarce. Many UDA methods for semantic segmentation leveraged adversarial learning frameworks~\cite{adv}. 
These adversarial approaches~\cite{adv1, adv2, adv3} involved training model to generate features that could deceive a discriminator, effectively aligning the feature distributions of source and target domains.
Recently, self-training approaches~\cite{self-training1, ProDA, ProCA} have emerged to incrementally improve the pseudo-labels.

While most existing works consider UDA between two domains of the same modality (e.g., RGB), MS-UDA{~\cite{MS-UDA}} proposed a UDA technique between the RGB-thermal domains to address the data scarcity issue in the thermal image dataset.
To solve this issue, MS-UDA proposed one-way knowledge distillation strategies to train the thermal image semantic segmentation: i) large-scale RGB-to-RGB inter-domain adaptation, ii) RGB-to-thermal, and iii) thermal-to-thermal intra-domain adaptation.
However, this distillation process simply transfers teacher predictions to the student without considering the characteristics of each domain, and the performance degradation for each distillation stage is significant. For effective UDA in cross-spectral domains, we design a novel UDA framework that leverages the complementary knowledge of the RGB-thermal domains by filtering out the inherent drawbacks associated with each spectral domain.

\section{METHOD}\label{Method}

\begin{figure*}[!t]
    \centering
    \includegraphics[width=0.85\linewidth]{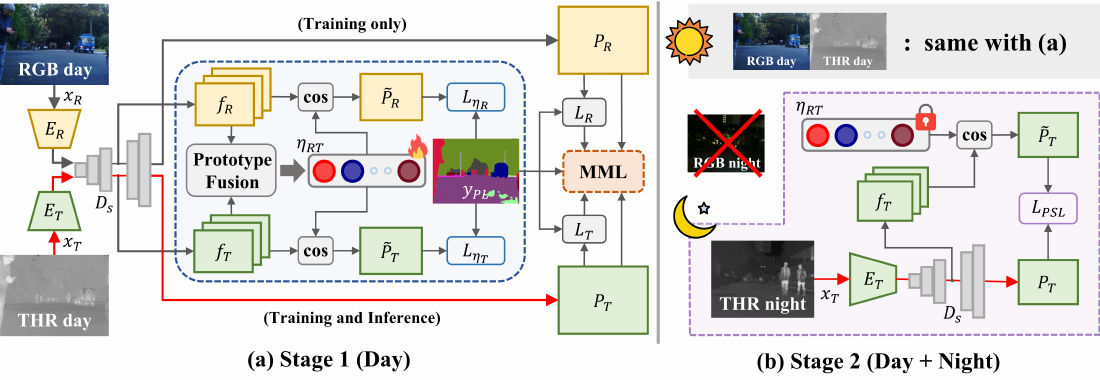}
        \caption{\label{overall} An overview of our framework. (a) In stage 1, both RGB and thermal networks are trained in the daytime using pseudo-labels $y_{PL}$ generated by HRNet~\cite{HRNet} pre-trained on a large-scale RGB dataset~\cite{CityScapes}. Simultaneously, Masked Mutual Learning (MML) is applied between these student networks, and cross-spectral prototypes $\eta_{RT}$ are gradually updated during training time. (b) In stage 2, the same learning process is performed for daytime as in (a). We impose prototypical self-supervised loss $L_{PSL}$ using our cross-spectral prototypes to address the absence of reliable annotations for nighttime training. We note that gray lines indicate processes performed only during the training phase, while red lines indicate processes in both the training and inference phases. The cos also refers to the cosine similarity function.}
    \label{fig:overall}
    \vspace*{-\baselineskip}
\end{figure*}

\begin{figure}[!t]
    \centering
    \includegraphics[width=0.9\linewidth]{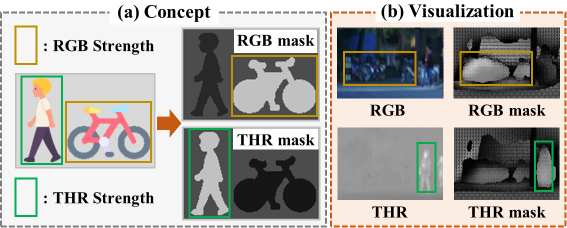}
    \caption{\label{MML} We present a conceptual illustration of masks in Masked Mutual Learning (MML) (a), and the visualization results (b). 
    (a) We generate RGB and THR masks by incorporating intra- and inter-spectral masks calculated from uncertainty maps for each spectral segmentation prediction. These masks exploit the strengths of each domain while mitigating its inherent limitations (e.g., person in RGB and bicycle in THR), providing complementary training signals to each spectral student network.
    (b) The visualization results of our masks during training.}
    \label{fig:fig3}
    \vspace*{-\baselineskip}
\end{figure}

\subsection{Overview}

\textbf{Overall Architecture.}
Our goal is to train the thermal image semantic segmentation network while mitigating the spectral domain discrepancies. To perform this, we introduce several enhancements that are employed exclusively during the training phase to ensure persistent efficiency during inference. 
As illustrated in Fig.~\ref{fig:overall}-(a), our method consists of three parts: spectral-specific encoder, weight-shared decoder, and prototype-fusion module. Specifically, an RGB image $x_{R}$ and a thermal image $x_{T}$ are fed into a spectral-specific encoder $E_{\theta}$, where $\theta = \{R, T\}$. The encoder $E_{\theta}$ extracts the multi-scale feature map and then each spectral feature map is passed through a weight-shared decoder $D_{s}$ with a skip connection between encoder and decoder features, producing the multi-scale decoder features and segmentation output. 
We then calculate cross-entropy loss $L_{seg}$ between the pseudo-labels $y_{PL}$ and the prediction of each spectral stream $P_{\theta}$ as follows: 
\begin{equation} \label{eq1}
\begin{split}
         L_{seg} &= L_R + L_T \quad \text{where} \\
         L_{\theta} &= -\frac{1}{N}\sum_{p=1}^{H\times W}\sum_{c=1}^{C} y_{PL}^{(p,c)} \log{(P_{\theta}^{(p,c)})}
\end{split}
\end{equation}
where H, W, N, and C denote the height, width, the number of pixels, and classes. Pseudo-labels $y_{PL}$ are generated by HRNet~\cite{HRNet} pre-trained on a large-scale RGB dataset~\cite{CityScapes}.

\textbf{Cross-Spectral Prototypes.}
Inspired by previous methods~\cite{ProDA, ProCA, Bi-directional}, our UDA framework leverages prototypes to incorporate semantic information at the pixel level.
First, we generate cross-spectral prototypes that share the latent space of two spectral domains. Specifically, these cross-spectral prototypes $\eta_{RT}$ are generated by using intermediate decoder features $f_{\theta} \in \mathbb{R}^{\frac{H}{4} \times \frac{W}{4}}$ as follows: 
\begin{equation} \label{eq2}
\begin{split}
        \eta_{\theta}^{(c)} =&  \frac{\sum_{p\in{\Omega}} f_{\theta}^{(p)} \odot \Idf[y_{PL}^{(p,c)}=1] } 
{\sum_{p\in{\Omega}}\Idf[y_{PL}^{(p,c)}=1]} \\
&\eta_{RT}^{(c)} = \frac{\eta_R^{(c)} + \eta_T^{(c)}}{2}
\end{split}
\end{equation}
where $\Omega$ denotes a set of pixels that has a higher confidence score than 0.1, $\odot$ is an element-wise multiplication, $\Idf[\cdot]$ is an indicator function, and $y_{PL}^{(p,c)}$ is one-hot labels, i.e., 1 if the class label at position $p$ corresponds to $c$ and 0 otherwise. These prototypes approximate the representational centroids shared by both spectral representations for each class.
We also use        the cross-spectral prototypes to perform contrastive learning for each spectral domain. Specifically, we encourage each pixel of the $f_{\theta}^{(p)}$ to be attracted toward prototypes of the same class while pushing away different classes. We calculate prototypical contrastive loss $L_{\eta}$ as follows:
\begin{equation} \label{eq3}
\begin{split}
        L_{\eta} &= L_{\eta_R} + L_{\eta_T} \quad \text{where} \\
       L_{\eta_\theta} &= -\sum_{p=1}^{H\times W}\sum_{c=1}^{C}y_{PL}^{(p,c)}\log\frac{\exp(s(f_{\theta}^{(p)},\eta_{RT}^{(c)})/\tau)}{\sum_c\exp(s(f_{\theta}^{(p)},\eta_{RT}^{(c)})/\tau)}
\end{split}
\end{equation}
where $\tau$ and $s(\cdot,\cdot)$ refer to the temperature and cosine similarity, respectively. $\tau$ is set to 1.
Cross-spectral prototypes are progressively updated using an exponential moving average with a momentum parameter of 0.9 during the training to capture richer semantic knowledge.

\subsection{Boosting Cross-Spectral UDA for Thermal Segmentation}

\textbf{Masked Mutual Learning.}
RGB and thermal cameras exhibit complementary strengths, with RGB sensors providing rich visual information such as color and texture, while thermal sensors facilitate robust perception in low-light environments. Based on this aspect, we propose Masked Mutual Learning (MML), which encourages spectral-specific mutual learners to share their advantageous knowledge. 
The main idea of MML is to maximize the complementary information between cross-spectral images during training by eliminating the inherent drawback within each spectral feature. We leverage each spectral-wise mask in the MML strategy, as depicted in Fig.~\ref{MML}-(a). The RGB network is guided by the output of the thermal network, focusing on regions where the thermal mask is activated. The thermal network is also guided in the same manner by the RGB mask.
To accomplish this, we design uncertainty-aware masking that eliminates uncertain training signals during mutual learning. Specifically, uncertainty-aware masks are calculated by incorporating intra-spectral and inter-spectral relationships.

To quantify uncertainty, we first calculate cross-entropy loss maps $L_{\theta} \in \mathcal{R}^{H \times W \times 1}$ for each spectral using student and teacher predictions. Subsequently, we concatenate the loss maps along the channel axis and apply softmax to generate an inter-spectral mask as follows:
\begin{equation}
\label{eq4}
\begin{split}
\mathcal{M}_{R}^{inter},\mathcal{M}_{T}^{inter}=(\text{split} \circ \text{softmax} \circ \text{cat})(-L_{R}, -L_{T}).
\end{split}
\end{equation}
The inter-spectral mask assigns weights to spectral prediction during mutual learning based on which spectral-loss map is larger than the counterparts.
Moreover, the intra-spectral uncertainty mask is generated by applying a sigmoid function $\sigma$ as follows: 
\begin{equation}
\label{eq:bihome_loss}
\begin{split}
\mathcal{M}_{R}^{intra} &= \alpha(1-\sigma(L_{R})),\; \mathcal{M}_{T}^{intra} = \alpha(1-\sigma(L_{T}))\\
\end{split}
\end{equation}
where $\alpha$ is set to 2. This mask can filter out inaccurate predictions even when they outperform the results of their counterparts. Finally, our masked mutual loss $L_{MML}$ is imposed via KL divergence loss with intra- and inter-spectral uncertainty masks as follows:
\begin{equation} \label{eq6}
\begin{split}
        L_{MML} &= L_{KL}(P_{R}M_{T},P_{T}M_{T}) \\
            &+ L_{KL}(P_{T}M_{R},P_{R}M_{R}) \\
            M_{\theta} &= \mathcal{M}_{\theta}^{intra} \odot \mathcal{M}_{\theta}^{inter}.
\end{split}\
\end{equation}

\textbf{Prototypical Self-Supervised Learning.}\label{stage2}
 Due to the absence of manual annotation for training, it is essential to obtain reliable pseudo-labels from the pre-trained network.
However, the susceptibility to domain shift and the inherent limitations of RGB sensors in low-light conditions can lead to inaccurate pseudo-labels, hindering the effective training of student networks. While MS-UDA employed CycleGAN~\cite{CycleGAN} to generate a daytime thermal image to nighttime synthetic image for training, it is limited by the model's inability to learn the distribution of real night conditions. 

To solve this, we propose a prototypical self-supervised loss \textbf{(PSL)} that employs the segmentation output of the decoder as pseudo-labels and transfers them to outputs of prototypes. This strategy promotes the model to learn the segmentation on the real night distribution despite the absence of manual annotations and the inherent limitations of pre-trained RGB models in low-light conditions. As illustrated in Fig.~\ref{fig:overall}-(b), the model is trained on both daytime and nighttime images during the second learning stage. The daytime learning process proceeds in the same manner as the initial learning process described in Fig.~\ref{fig:overall}-(a). In the night condition, a night thermal image is fed into the thermal encoder and decoder, resulting in segmentation outputs from the decoder and the prototypes, respectively. The decoder's output is then employed as a pseudo-label, and the KL divergence loss is imposed as follows:
\begin{equation} \label{eq7}
\begin{split}
        L_{PSL} = L_{KL}(\Tilde{P}_T,P_T) \\
\end{split}
\end{equation}
where $\Tilde{P}_T$ and $P_T$ refer to segmentation outputs generated by the prototypes and decoder, respectively.

Since our cross-spectral prototypes can be updated using RGB and thermal features as shown in Eq.~(\ref{eq2}), we only selectively update our cross-spectral prototypes in the daytime training set and freeze the prototypes when the model is trained on nighttime-condition images.

\subsection{Training Loss}

\textbf{Stage 1.} 
As shown in Fig.~\ref{fig:overall}-(a), stage 1 of our proposed framework leverages daytime images for training. We define a training loss as follows:
\begin{equation}
\label{loss_stage1}
    \begin{split}
    L_{stage1} = L_{seg} + \lambda_{1}L_{\eta} + \lambda_{2}L_{MML}
    \end{split}
\end{equation}
where $\lambda_{1}$ and $\lambda_{2}$ are set to 0.2, 20, respectively.

\textbf{Stage 2.}
In stage 2, we impose a PSL for nighttime in the absence of reliable pseudo-labels, as depicted in Fig.~\ref{fig:overall}-(b). The same process is performed for daytime images, as in Fig.~\ref{fig:overall}-(a). We calculate the training loss for stage 2 as follows:
\begin{equation}
\label{loss_stage2}
    \begin{split}
    L_{stage2} = L_{stage1} + \lambda_{3}L_{PSL}\\
    \end{split}
\end{equation}
where $\lambda_{3}$ is set to 20. 
$L_{stage1}$ is computed using daytime cross-spectral images, while $L_{PSL}$ is calculated from nighttime single thermal image.

\section{EXPERIMENTS}
\subsection{Implementation Details} \label{Imple_Details}

\textbf{Dataset and Evaluation Metric.} For training our framework, we generate pseudo-labels with a teacher network~\cite{HRNet} trained on Cityscapes~\cite{CityScapes}, a large-scale RGB dataset (i.e., 5,000 images) that consists of 19 classes, the same setting in MS-UDA~\cite{MS-UDA}.  
We conduct our experiments on two public RGB-T paired datasets: the MF dataset~\cite{MFNet} and the KAIST Multispectral Pedestrian Detection (KP) dataset~\cite{KP}. 

As the MF dataset is annotated in 9 classes, we only report the mean intersection over union (mIoU) on the three classes that overlap with Cityscapes (i.e., car, person, and bicycle).
In contrast to mixed day-night evaluation protocols in previous methods, MS-UDA exclusively trains on 820 daytime images and tests on 749 nighttime images.
Due to this setting, it is difficult to make fair comparisons with other previous methods.
Therefore, we re-implemented MS-UDA with 410 training daytime images and marked it as MS-UDA* in the experiment table. All the evaluations were conducted on 393 day-night testing images. 

The KP dataset consists of RGB-T paired 95K video frames (62.5K for daytime and 32.5K for nighttime) on the urban driving scene for pedestrian detection.
For semantic segmentation, MS-UDA manually annotated 950 images with the same 19 class ground-truth labels as Cityscapes.
Recently, Shin \textit{et al.}~\cite{CRM_Seg} divided 950 annotated images into 499, 140, and 311 for training, validation, and testing, respectively, and we followed their setting.

\textbf{Training Details.} As in MS-UDA~\cite{MS-UDA}, we adopt basic structure of RTFNet~\cite{RTFNet} consists of an encoder-decoder network. We also employed ImageNet~\cite{ImageNet} pre-trained ResNet50 as an encoder. 
All experiments were conducted with a batch size of 8 on 2 NVIDIA GeForce RTX 3090 GPUs. 

At the initial learning stage (i.e., Ours w/o stage 2 on the experiment tables), we train the model with 120$k$ iterations for the MF dataset and 500$k$ iterations for the KP dataset. We adopt the same training data split as MS-UDA, utilizing 410 unlabeled daytime images for training the MF dataset and 3,283 unlabeled daytime images for training the KP dataset. We also utilize fake night thermal images generated from CycleGAN~\cite{CycleGAN} with a 50\% probability.
In the case of the second learning stage, we train the model using only 30\% of the iterations trained in stage 1. We utilize 374 and 3,095 additional unannotated nighttime images for training and fake night images used in stage 1 are not employed. 

\begin{table}[t]
    \centering
    \caption{Quantitative comparison with state-of-the-art methods on MF day-night evaluation set~\cite{MFNet}. 
    (D+N: training with daytime and nighttime images
    ; 
    R+T: testing with RGB-T image pairs)
    } 
    \label{table:MF}
    \begin{adjustbox}{width=\linewidth}
        \begin{tabular}{c c c c| c c c c}
            \toprule
            & Method & Train & Test & Car & Person & Bicycle & \textbf{mIoU$\uparrow$} \\
            \midrule
            
            \multirow{7}{*}{\underline{Sup.}} & MFNet~\cite{MFNet}& D+N & R+T  & 65.9 & 58.9 & 42.9 & 55.9\\   
            & RTFNet~\cite{RTFNet} &D+N&  R+T & 86.3 & 67.8 & 58.2 & 70.7\\  
            & MDBFNet~\cite{MDBFNet} &D+N&  R+T & 85.9 & 69.2 & 58.9 & 71.3 \\
            & CENet~\cite{CEKD} &D+N&  R+T & 85.8 & 70.0 & 61.4 & 72.4 \\  
            & EAEFNet~\cite{EAEFNet} &D+N&  R+T & 86.8 & 71.8 & 62.0 & 73.5 \\  
            & CMX~\cite{CMX} &D+N&  R+T & \underline{90.1} & \underline{75.2} & 64.5 & 76.6 \\ 
            & CRM~\cite{CRM_Seg} &D+N & R+T & 90.0 & 75.1 & \underline{67.0} & \underline{77.4} \\ 
            \midrule

            \multirow{9}{*}{\textbf{UDA}} & ProCA~\cite{ProCA} & D & T & 50.8 & 36.8 & 14.2 & 33.9 \\
            & DAFormer~\cite{DAFormer} & D& T & 52.0 & 51.6 & 38.9 & 47.5 \\
            & MS-UDA*~\cite{MS-UDA} &  D & T & 82.1 & 73.4 & 55.6 & 70.4 \\  
            & \ccgray Ours w/o stage 2 & \ccgray D & \ccgray T &  \ccgray 85.7 & \ccgray 78.4 & \ccgray 61.0 & \ccgray 75.0 \\  
            \cmidrule(){2-8}
            
             & ProCA~\cite{ProCA} & D+N & T & 48.9 & 47.1 & 15.4 & 37.1 \\
            & DAFormer~\cite{DAFormer} & D+N& T & 48.9 & 55.4 & 28.9 & 44.4\\
            &HeatNet~\cite{HeatNet} & D+N& R+T & 56.4 & 68.8 & 33.9 & 53.0 \\
            & EKNet~\cite{CEKD} & D+N& T & 78.6 & 67.5 & 51.9 &  66.0 \\ 
            & \ccgray Ours &\ccgray  D+N & \ccgray T & \ccgray \textbf{86.2} & \ccgray \textbf{80.3} & \ccgray \textbf{64.3} & \ccgray \textbf{76.9} \\  
    
            \bottomrule
        \end{tabular}
    \end{adjustbox}  
    \begin{tablenotes}
        \item [] \scriptsize{* We re-implemented MS-UDA with 410 training daytime images}
    \end{tablenotes}
    \vspace*{-\baselineskip}
\end{table}

\subsection{Main Results} 
\textbf{Quantitative Results on the MF Dataset~\cite{MFNet}.} Table~\ref{table:MF} demonstrates the effectiveness of our framework on the MF dataset.
To compare the effectiveness of each learning strategy, the networks for all UDA methods except HeatNet~\cite{HeatNet} are set to RTFNet~\cite{RTFNet} with the ResNet50-based encoder. 

Compared with existing methods, our UDA framework outperforms most of the previously supervised and all UDA methods. Furthermore, we achieve comparable performance to the state-of-the-art supervised method~\cite{CRM_Seg} only with a single thermal image as input. Specifically, ProCA~\cite{ProCA} and DAFormer~\cite{DAFormer} achieve poor results in cross-spectral images despite demonstrating superior performance in the RGB domain. These methods indicate a substantial mIoU drop for specific classes, especially the bicycle class in thermal images, due to inherent deficiencies of thermal cameras (e.g., difficulty in capturing objects with low thermal emissivity).
These results imply the necessity for a cross-spectral UDA strategy that can effectively leverage the complementary knowledge of the RGB and thermal domains. On the other hand, our method achieves superior performance compared to existing state-of-the-art approaches in single-spectral~\cite{ProCA}, \cite{DAFormer} and cross-spectral domains~\cite{HeatNet}, \cite{MS-UDA}, \cite{CEKD} even when employing only stage 1 (i.e., Ours w/o stage 2).
Moreover, our final model (i.e., Ours) exhibits a remarkable improvement in mIoU of 6.5 over the baseline~\cite{MS-UDA}, demonstrating the effect of our PSL in enabling segmentation training without manual annotations or a pre-trained teacher network.

\begin{table*}[t]
    \centering
    \caption{Quantitative comparison with state-of-the-art methods on KP day-night evaluation set~\cite{KP}.
    (D+N: training with daytime and nighttime images
    ; R+T: testing with RGB-T image pairs
    } 
    \label{table:KP}
    \begin{adjustbox}{width=\linewidth}
        \begin{tabular}{c c c c | c c c c c c c c c c c c c c c c c c c c}
            \toprule
             &Method&Train&Test&\rotatebox{90}{Road}&\rotatebox{90}{Sidewalk}&\rotatebox{90}{Building}&\rotatebox{90}{Wall}&\rotatebox{90}{Fence}&\rotatebox{90}{Pole}&\rotatebox{90}{Traffic light}&\rotatebox{90}{Traffic sign}&\rotatebox{90}{Vegetation}&\rotatebox{90}{Terrain}&\rotatebox{90}{Sky}&\rotatebox{90}{Person}&\rotatebox{90}{Rider}&\rotatebox{90}{Car}&\rotatebox{90}{Truck}&\rotatebox{90}{Bus}&\rotatebox{90}{Train}&\rotatebox{90}{Motorcycle}&\rotatebox{90}{Bicycle}&\textbf{mIoU$\uparrow$}\\
            \midrule
            \multirow{4}{*}{\underline{Sup.}} &MFNet~\cite{MFNet}&D+N&R+T&93.5&23.6&75.1&0.0&0.1&9.1&0.0&0.0&69.3&0.2&90.4&24.0&0.0&69.6&0.3&0.3&0.0&0.0&0.6&24.0\\
            
            &RTFNet~\cite{RTFNet}&D+N&R+T&94.6&39.4&86.6&0.0&0.6&0.0&0.0&0.0&81.7&3.7&92.8&58.4&0.0&87.7&0.0&0.0&0.0&0.0&0.5&28.7\\

            &CMX~\cite{CMX}&D+N&R+T&97.7&53.8&90.2&0.0&47.1&46.2&10.9&45.1&87.2&\underline{34.3}&93.5&74.5&0.0&91.6&0.0&59.7&0.0&46.1&0.2&46.2\\

            &CRM~\cite{CRM_Seg}&D+N&R+T&\underline{99.0}&\underline{61.9}&\underline{91.8}&0.0&\underline{58.7}&\underline{50.6}&\underline{39.2}&\underline{55.3}&\underline{89.2}&23.2&\underline{94.3}&\underline{85.2}&\underline{2.9}&\underline{95.3}&0.0&\underline{80.5}&0.0&\underline{66.2}&\underline{54.6}&\underline{55.2}\\
            \midrule

            \multirow{7}{*}{\textbf{UDA}} & DAFormer~\cite{DAFormer} &D&T&67.5&0.1&0.4&0.0&0.1&1.9&0.0&3.6&22.3&0.4&4.3&0.0&0.0&0.0&0.0&0.0&0.0&0.0&0.0&5.3\\

            & ProCA~\cite{ProCA} &D&T&71.4&5.0&44.3&0.0&2.0&2.1&0.0&1.3&26.2&2.2&15.6&16.8&0.0&40.8&0.0&0.1&0.0&0.1&0.0&12.0\\

            & MS-UDA~\cite{MS-UDA} &D&T&97.2&25.7&86.3&0.0&31.4&16.5&0.0&28.5&83.1&25.1&92.5&60.8&0.0&85.1&0.0&78.3&0.0&0.0&0.0&37.4\\
            
             & \ccgray Ours w/o stage 2 &\ccgray D&\ccgray T&\ccgray 97.7&\ccgray 33.0&\ccgray \textbf{88.7}&\ccgray 0.0&\ccgray 35.5&\ccgray \textbf{36.4}&\ccgray 14.9&\ccgray 46.2&\ccgray 85.6&\ccgray \textbf{24.5}&\ccgray \textbf{94.4}&\ccgray \textbf{72.2}&\ccgray 4.9&\ccgray 88.4&\ccgray 0.0&\ccgray \textbf{82.9}&\ccgray 0.0&\ccgray 35.3&\ccgray \textbf{4.3}&\ccgray 44.5\\ 

            \cmidrule{2-24}
            & DAFormer~\cite{DAFormer} &D+N&T&57.6&0.1&0.0&0.0&0.0&1.0&0.0&4.9&28.8&3.5&19.1&0.0&0.0&0.0&0.0&0.0&0.0&0.0&0.0&6.1\\

            & ProCA~\cite{ProCA} &D+N&T&74.0&5.3&48.8&0.0&1.4&2.3&0.0&0.3&21.1&2.3&11.6&17.9&0.0&22.5&0.0&0.3&0.0&0.0&0.0&10.9\\
            
            &\ccgray Ours &\ccgray D+N&\ccgray T&\ccgray \textbf{97.8}&\ccgray \textbf{34.4}&\ccgray 88.6&\ccgray 0.0&\ccgray \textbf{36.6}&\ccgray 35.0&\ccgray \textbf{17.9}&\ccgray \textbf{47.3}&\ccgray \textbf{85.7}&\ccgray \textbf{24.5}&\ccgray \textbf{94.4}&\ccgray 71.9&\ccgray \textbf{5.5}&\ccgray \textbf{89.3}&\ccgray \textbf{0.2}&\ccgray 81.7&\ccgray 0.0&\ccgray \textbf{56.0}&\ccgray 2.6&\ccgray \textbf{45.8}\\ 
            
            \bottomrule
        \end{tabular}
    \end{adjustbox}  
    \vspace*{-\baselineskip}
\end{table*}

\begin{figure*}[!t]
    \centering
    \includegraphics[width=\linewidth]{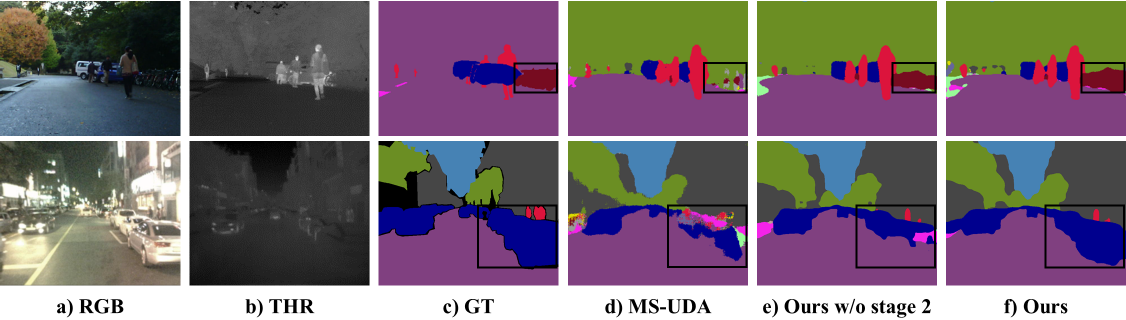}
    \caption{ Qualitative comparison with MS-UDA~\cite{MS-UDA} on the MF~\cite{MFNet} and KP~\cite{KP} datasets. We visualize the prediction result of the daytime image for the MF dataset (first row) and the nighttime image for the KP dataset (second row). In comparison to MS-UDA, our approach shows robust performance for both daytime and nighttime across all classes.}
    \label{fig:result}
    \vspace*{-\baselineskip}
\end{figure*}

\textbf{Quantitative Results on the KP Dataset~\cite{KP}.}
In Table II, we present the performance on the KP dataset. Consistent with the observations on the MF dataset, the state-of-the-art methods for single and cross-spectral UDA show insufficient segmentation results since they do not use the benefits of mutual information between cross-spectral images during distillation. However, our method significantly surpasses the advanced cross-spectral UDA method~\cite{MS-UDA}, despite employing the same model as in ~\cite{MS-UDA} (37.4 vs 44.5 in mIoU).
Thermal images at night exhibit low contrast due to the reduced heat variation compared to daytime. This can reduce the quality of nighttime pseudo-labels generated from prototypes, potentially leading to lower mIoU in a few classes (e.g., Pole and Bus) with stage 2 training. Nevertheless, imposing the proposed PSL in stage 2 ultimately achieved 45.8 mIoU by improving performance in most classes.

\textbf{Qualitative Results.} 
Fig.~\ref{fig:result} shows qualitative comparisons on the MF and KP datasets.  
As shown in the black box of row 1, MS-UDA inaccurately predicts the bicycle class. In contrast, our proposed method provides successful prediction when only employing MML and cross-spectral prototypes (i.e., Ours w/o stage 2).
These results are attributed to our masked mutual learning that encourages the student models to effectively transfer complementary knowledge within cross-spectral images to each other by filtering out the inherent limitations associated with each spectral domain.
Moreover, our model makes good predictions at night compared to MS-UDA, as shown in row 2. These results show that our PSL with the cross-spectral prototype effectively enhances performance in night conditions, where RGB pre-trained networks struggle with reliable annotation generation.

\subsection{Ablation Study}

\textbf{One-Way Distillation vs Mutual Learning.}  In Table~\ref{table:ablation}, we present an ablation analysis for each component of our framework on the MF dataset. 
We first replace the sequential one-way distillation in MS-UDA with a two-way distillation framework. 
To achieve this, we train the student models by imposing $L_{seg}$ and $L_{MML}$ without applying our domain-wise masks $M_{R}$ and $M_{T}$.
Interestingly, this two-way distillation method achieves remarkable mIoU gains for both thermal (70.4 vs 72.8) and RGB segmentation (61.7 vs 69.4). This suggests that complementary information between cross-spectral domains is important.

\begin{table}[t]
    \centering
    \caption{Ablation results for each masking strategy applied to Masked Mutual Learning (MML) on MF dataset~\cite{MFNet}}
    \label{table:mask}
    \begin{adjustbox}{width=0.7\linewidth}
        \begin{tabular}{c | c c c c}
            \toprule
            \makecell{Masking \\ Strategy}
            & Car & Person & Bicycle & \textbf{mIoU$\uparrow$} \\
            \midrule
            w/o Mask & 82.7 & 76.7 & 58.9 & 72.8 \\
            Intra & \textbf{85.3} & \textbf{78.0} & 56.1 & 73.1 \\
            Inter & 83.9 & 77.8 & 57.2 & 72.9 \\
            Intra \& Inter & 85.0 & 76.6 & \textbf{59.0} & \textbf{73.6} \\
            \bottomrule
        \end{tabular}
    \end{adjustbox}  
\end{table}

\begin{table}[t]    
    \centering
    \caption{Ablation results of components in our proposed framework. We report the mIoU on MF dataset~\cite{MFNet}}
    \label{table:ablation}
    \begin{adjustbox}{width=0.85\linewidth}
        \begin{tabular}{c c c | c c c | c}
            \toprule
            \multirow{2}{*}[-.3em]{MML} & \multirow{2}{*}[-.3em]{\makecell{Cross-spectral \\ Prototypes}} 
            & \multirow{2}{*}[-.3em]{Stage 2} & \multicolumn{3}{c|}{Thermal} & RGB \\
            \cmidrule(){4-7}
            & & & Day & Night& \textbf{All} & \textbf{All} \\
            \midrule
            \multicolumn{3}{c|}{One-way distillation~\cite{MS-UDA}} & 71.6 & 67.0 & 70.4 & 61.7 \\
            \midrule
            \multicolumn{3}{c|}{Two-way distillation}& 75.3&67.9&72.8&69.4  \\
            \checkmark & & & 76.4&67.4&73.6&69.2 \\
            \checkmark & \checkmark & & 77.2&70.5&75.0&71.1 \\
            \checkmark & \checkmark & \checkmark & \textbf{78.8} & \textbf{73.2} & \textbf{76.9} & \textbf{73.4}  \\
            
            \bottomrule
        \end{tabular}
    \end{adjustbox}  
    \vspace*{-\baselineskip}
\end{table}

\textbf{The Effectiveness of Masked Mutual Learning.} Although mutual learning facilitates efficient model learning between students, it may also lead to the dissemination of potentially unreliable knowledge. To address this issue, our masked mutual learning can be employed to refine the knowledge by considering both inter and intra-spectral dependencies. 
To validate the effect of MML, we ablate the uncertain-aware masking components. In Table~\ref{table:mask}, our intra and inter-spectral masks bring minor performance gains when applied independently (72.8 $\rightarrow$ 72.9, 73.1). However, when they are combined, we achieve a significant performance gain compared to the mutual learning model without a mask (72.8 $\rightarrow$ 73.6 in mIoU). This implies that it is crucial to consider both the intra and inter-spectral components for calculating uncertain masks.

\textbf{The Benefits of Cross-Spectral Prototypes.} We also evaluated the cross-spectral prototypes in stage 1 and achieved an impressive improvement of mIoU in both thermal (73.6 $\rightarrow$ 75.0) and RGB domains (69.2 $\rightarrow$ 71.1) as shown in the fourth row of Table~\ref{table:ablation}. This implies that our method effectively learns scant semantic information (e.g., a person in the RGB domain and a bicycle in the thermal domain) inherently limited to individual domains by capturing complementary knowledge between spectral domains through cross-spectral prototypes.

\textbf{The Effectiveness of Prototypical Self-Supervised Loss.} As mentioned in Section~\ref{Method}, we designed a PSL in stage 2 for self-supervision at nighttime due to the absence of reliable pseudo-labels. In Table~\ref{table:ablation}, we observe that our PSL leads to a significant performance boost in mIoU for the day (77.2 $\rightarrow$ 78.8) and night times (70.5 $\rightarrow$ 73.2). Interestingly, despite that our PSL is only imposed on a thermal student network, the RGB student model also achieves meaningful improvement (71.1 $\rightarrow$ 73.4). These results suggest that PSL is effective in training the model despite the absence of reliable ground truth labels at nighttime.

\textbf{Generalizability Across Diverse Segmentation Networks.} Consistent with MS-UDA~\cite{MS-UDA}, we leverage RTFNet~\cite{RTFNet} as our segmentation network. To assess the generalizability of our proposed learning strategy, we conduct an experiment utilizing DeepLab-V3~\cite{DeepV3}, a widely employed network architecture for segmentation tasks. As shown in Table~\ref{table:backbone}, our method outperforms MS-UDA across both RTFNet and DeepLab-V3 architectures, highlighting its generalizability and robustness to model architectures.

\begin{table}[t]
    \centering
    \caption{Comparison of generalization performance across diverse segmentation network architectures on MF dataset~\cite{MFNet}}
    \label{table:backbone}
    \begin{adjustbox}{width=0.9\linewidth}
        \begin{tabular}{c c | c c c c}
            \toprule
            Network & Method & Car & Person & Bicycle & \textbf{mIoU$\uparrow$} \\
            \midrule
            
            \multirow{3}{*}{RTFNet~\cite{RTFNet}} & MS-UDA*~\cite{MS-UDA} & 82.1 & 73.4 & 55.6 & 70.4  \\
            & Ours w/o stage 2 & 85.7 & 78.4 &61.0 &75.0   \\
            & Ours & \textbf{86.2} & \textbf{80.3} & \textbf{64.3} & \textbf{76.9}   \\
            \midrule
            
            \multirow{3}{*}{DeepLab-V3~\cite{DeepV3}} & MS-UDA*~\cite{MS-UDA} & 77.4 & 74.7 & 46.4 & 66.2  \\
            & Ours w/o stage 2 & 82.2 & 75.6 & 50.1 & 69.3 \\
            & Ours & \textbf{82.6} & \textbf{75.7} & \textbf{53.0} & \textbf{70.5}   \\
            
            \bottomrule
        \end{tabular}
    \end{adjustbox}  
    \begin{tablenotes}
        \item [] \scriptsize{* We re-implemented MS-UDA with 410 training daytime images}
    \end{tablenotes}
    \vspace*{-\baselineskip}
\end{table}

\section{CONCLUSIONS}

In this letter, we present a cross-spectral unsupervised domain adaptation (UDA) approach for thermal image semantic segmentation. Our proposed Masked Mutual Learning (MML) strategy facilitates effective UDA by enabling the transfer of essential information between RGB and thermal domains. Moreover, we introduce cross-spectral prototypes to incorporate pixel-wise semantic knowledge. These prototypes are subsequently employed within a novel prototypical self-supervised loss function, enabling robust training even under unreliable nighttime conditions. Experimental results demonstrate the effectiveness of our framework, which significantly outperforms previous UDA methods while achieving competitive results with state-of-the-art supervised methods. In the future, we will investigate novel pseudo-label enhancement methodologies to refine our framework.










\end{document}